\documentclass{article}

\usepackage[dvipsnames]{xcolor}
\usepackage{arxiv}

\usepackage[english]{babel}
\usepackage[utf8]{inputenc} 
\usepackage[T1]{fontenc}    
\usepackage{hyperref}       
\usepackage{xcolor}
\hypersetup{
    colorlinks=true,
    linkcolor=blue,
    filecolor=magenta,      
    urlcolor=cyan,
}
\usepackage{url}            
\usepackage{booktabs}       
\usepackage{amsfonts}       
\usepackage{nicefrac}       
\usepackage{microtype}      
\usepackage{lipsum}
\usepackage{graphicx}
\graphicspath{ {./images/} }
\usepackage[noend]{algpseudocode}
\usepackage{amsmath, amsfonts, amsthm, mathtools, amssymb, amsthm}
\usepackage{tikz}
\usepackage{caption}
\usepackage{float}
\usepackage{subfiles}
\usepackage{subfig}
\usepackage{array}
\usepackage[ruled,linesnumbered]{algorithm2e}

\usepackage{biblatex}
\usepackage{multirow}
\usepackage{multicol}
\usepackage{booktabs}
\usepackage{siunitx}
\usepackage{amsmath}
\usepackage{kotex}
\definecolor{bluebell}{rgb}{0.39, 0.68, 0.88}
\definecolor{caribbeangreen}{rgb}{0.29, 0.78, 0.54}
\DeclareMathOperator*{\argmin}{argmin}  

\title{Quantization of Generative Adversarial Networks for Efficient Inference: a methodological study}

\author{
Pavel Andreev \\
 
   Higher School of Economics \\
  Skolkovo Institute of Science and Technology \\
 Samsung AI Center Moscow\\
  Moscow, Russia \\
  \texttt{Pavel.Andreev@skoltech.ru} \\
     \And
Alexander Fritzler\\
   Higher School of Economics \\
  Skolkovo Institute of Science and Technology \\
 Yandex
 \\
  Moscow, Russia \\
  \texttt{afritzler449@gmail.com} \\
       \And
Dmitry Vetrov\\
Samsung-HSE Laboratory \\
   Higher School of Economics \\
 Samsung AI Center Moscow\\
  Moscow, Russia \\
  \texttt{vetrovd@yandex.ru} \\
}

\begin{document}
\maketitle
\begin{abstract}


\textit{Generative adversarial networks} (GANs) have an enormous potential impact on digital content creation, e.g., photo-realistic digital avatars, semantic content editing, and quality enhancement of speech and images.
However, the performance of modern GANs comes together with massive amounts of computations performed during the inference and high energy consumption. 
That complicates, or even makes impossible, their deployment on edge devices. 
The problem can be reduced with \textit{quantization}---a neural network compression technique that facilitates hardware-friendly inference by replacing floating-point computations with low-bit integer ones.
While quantization is well established for discriminative models, the performance of modern quantization techniques in application to GANs remains unclear. GANs generate content of a more complex structure than discriminative models, and thus quantization of GANs is significantly more challenging.
To tackle this problem, we perform an extensive experimental study of state-of-art quantization techniques on three diverse GAN architectures, namely StyleGAN, Self-Attention GAN, and CycleGAN. 
As a result, we discovered practical recipes that allowed us to successfully quantize these models for inference with 4/8-bit weights and 8-bit activations while preserving the quality of the original full-precision models.

\textbf{\textit{Keywords:}} quantization, generative adversarial networks, compression
\end{abstract}

\section{Introduction}
	
	\label{ch:intro}

Generative adversarial networks (GANs) are the tool of choice for a variety of computer vision tasks involving generative modeling \cite{karras2019style, zhang2017beyond, zhu2017unpaired}. However, the expressiveness of modern deep neural networks comes along with tremendous computational and memory resources spent during the inference phase, with GANs being a clear illustration of this tendency \cite{brock2018large}. This circumstance significantly complicates the deployment of such models in real-world applications, especially on edge devices, where memory and latency are of the main concern. 

Over the past few years, considerable research attention has been dedicated to the compression of deep neural networks. The most popular compression techniques include pruning \cite{he2017channel, molchanov2017variational}, efficient architecture design \cite{chollet2017xception, howard2017mobilenets}, knowledge distillation \cite{hinton2015distilling, chen2017learning}, and quantization \cite{krishnamoorthi2018quantizing, yao2020hawqv3}. The last one is of particular research attention due to several reasons. First, it is “orthogonal” to the majority of other compression techniques, meaning that quantization can be applied to an already architecturally compressed and pruned model.
Second, the energy consumption of quantized models is decreased significantly due to reduced data movement \cite{krishnamoorthi2018quantizing}. And lastly, quantization facilitates the deployment of neural networks on FPU-free devices, as it enables integer-only inference \cite{yao2020hawqv3}.

Most of the relevant works explore quantization techniques in the context of image classification, whereas quantization of generative models is considered as a more challenging and understudied task \cite{wan2020deep}. In this paper, we reduce the concerned research gap by extensively studying the applicability of different quantization methods to generative adversarial networks. The key contributions of this work are as follows:
\begin{enumerate}
\item We explore the effectiveness of state-of-art post-training quantization techniques for quantization of style-based generative adversarial network. 
\item We carry out an experimental study of quantization-aware training techniques in application to three diverse GAN architectures, namely self-attention GAN \cite{zhang2019self}, cycle-consistent GAN \cite{zhu2017unpaired} and style-based GAN \cite{karras2019style}, and for the first time report successful 4-bit quantization of these models.
\item To the best of our knowledge, this is a pioneering work primarily dedicated to uniform quantization of different types of generative adversarial networks, providing a valuable starting point for future research. 
\end{enumerate}

\section{Related work}
	
	\label{ch:related}

\subsection{Quantization methods}

Quantization of neural networks is a popular compression technique that is based on the replacement of accurate floating-point operations by less accurate low-bit quantized operations. There have been significant advances in the quantization of deep classification models recently. The emerged methods can be roughly divided into post-training quantization (PTQ) and quantization-aware training (QAT) categories. Post-training quantization aims to quantize neural networks using a small part of the dataset (in some cases no data at all) for calibration of quantization parameters to ensure a certain local criterion (e.g., correspondence of minimum and maximum, MSE minimality). Recent work \cite{nagel2020up} showed that
minimizing the mean squared error (MSE) introduced in
the preactivations might be considered (under certain assumptions) as the best possible local criterion and performed optimization of rounding policy based on it. Works \cite{hubara2020improving} and \cite{wang2020towards} utilize the same local criterion but optimize weights and quantization parameters directly and employ per-channel weight quantization, thus considering a simplified task. BRECQ technique \cite{li2021brecq} further generalizes this idea by utilizing improved local objective enhanced with squared gradient information and taking advantage of block-wise granularity of the neural networks. In our experiments with PTQ, we use a similar approach, minimizing the MSE introduced in activations at the block-level granularity. Although post-training quantization of discriminative models has shown significant promise, it is still inferior to quantization-aware training, a technique that tunes model parameters directly on the task of interest using stochastic gradient descent. Despite QAT was originally used for weights fine-tuning only, recent works \cite{esser2019learned, jain2019trained} proposed to learn quantization parameters jointly with weights. The concerned methods employ the straight-through estimator (STE) technique to effectively differentiate through rounding operation, providing the state-of-art quantization quality.  The performance of the most noticeable PTQ and QAT techniques is summarized in Tables \ref{table_pqt} and \ref{table_qat}, respectively. In our work, we make use of the arguably most effective QAT technique, namely learned step size (LSQ \cite{esser2019learned}).

\begin{table}[h!]
\caption{Post-training quantization techniques. The table demonstrates ImageNet top1 accuracies for different architectures (RN -- ResNet, MBv2 -- MobileNetv2). The signed numbers in brackets $(\cdot)$ shows difference with the corresponding full-precision baseline reported in the paper. “wxay” means x-bit weights 
and y-bit activations.} \label{table_pqt}
\vskip 0.15in
\begin{center}
 \begin{tabular}{|>{\centering\arraybackslash}p{1.5cm}|>{\centering\arraybackslash}p{2cm}|>{\centering\arraybackslash}p{1.2cm}|>{\centering\arraybackslash}p{1.2cm}|>{\centering\arraybackslash}p{1.2cm}|>{\centering\arraybackslash}p{1.2cm}|}
\hline
  \multirow{2}{*}{Method} & \multirow{2}{*}{Per-channel}  & RN18 (a8/w8) & MBv2 (a8/w8) &  RN18 (a4/w4) &  RN50 (a4/w4) \\ 
  \hline
     Baseline \cite{krishnamoorthi2018quantizing} & \multirow{2}{*}{$\times$} & $69.2$ ($-0.5$)   & $0.1$ ($-71.8$) & \multirow{2}{*}{$-$} & \multirow{2}{*}{$-$}  \\ \hline
      Baseline \cite{krishnamoorthi2018quantizing} & \multirow{2}{*}{$\checkmark$} &  $69.7$ ($-0.1$)  & $69.7$ ($-2.2$) & \multirow{2}{*}{$-$} & \multirow{2}{*}{$-$}\\ \hline
   AdaRound \cite{nagel2020up} & \multirow{2}{*}{$\times$} &  $69.7$ ($+0$)  & $71.2$ ($-0.5$) & \multirow{2}{*}{$-$} & \multirow{2}{*}{$-$}\\ \hline
      CalibTIB \cite{hubara2020improving} & \multirow{2}{*}{$\checkmark$} & \multirow{2}{*}{$-$} & $73.0$ ($+0$) & $69.4$ ($-2.6$)  & $75.1$ $(-2.1)$ \\ \hline
    Bit-Split \cite{wang2020towards} & \multirow{2}{*}{$\checkmark$} & $69.7$ ($+0$) &  \multirow{2}{*}{$-$} & $67.5$ ($-2.1$) & $73.7$ ($-2.4$) \\ \hline 
    BRECQ \cite{li2021brecq} & \multirow{2}{*}{$\times$} & $71.0$ ($-0.1$)  & $72.4$ ($-0.1$) & $69.6$ ($-1.5$)  & $75.1$ ($-1.9$) \\ \hline     
\end{tabular}

\end{center}
\end{table}

\begin{table}[h!]
  \caption{Quatization-aware training techniques.  The table demonstrates ImageNet top1 accuracies for different architectures. Notations follow the table \ref{table_pqt} .} \label{table_qat}
  \vskip 0.15in
\begin{center}
    
 \begin{tabular}{|>{\centering\arraybackslash}p{1.5cm}|>{\centering\arraybackslash}p{2cm}|>{\centering\arraybackslash}p{1.2cm}|>{\centering\arraybackslash}p{1.5cm}|>{\centering\arraybackslash}p{1.2cm}|>{\centering\arraybackslash}p{1.2cm}|}
\hline
 \multirow{2}{*}{Method} & \multirow{2}{*}{Per-channel} & RN50 (a8/w8) & RN50 (a8/w4) & MBv2 (a8/w8) &  RN18 (a4/w4) \\ 
  \hline
    Baseline \cite{krishnamoorthi2018quantizing} & \multirow{2}{*}{$\checkmark$} & 75.0 ($-0.6$) & 73.0 ($-2.6$) & $70.9$ ($-1.0$) & \multirow{2}{*}{$-$} \\ \hline 
    \multirow{3}{*}{LSQ \cite{esser2019learned}} & \multirow{3}{*}{$\times$} & $76.8$ ($-0.1$) & $76.7$ ($-0.2$)  \textbf{a4/w4}  & \multirow{3}{*}{$-$} & $71.1$ ($+0.6$) \\ \hline 
    \multirow{2}{*}{TQT \cite{jain2019trained}} & $\times$  & $75.4$ ($+0$) & $74.4$ ($-1.0$)  & $71.8$ ($-0.1$) & \multirow{2}{*}{$-$} \\ \hline 
\end{tabular}

\end{center}
\end{table}

\subsection{Generative Adversarial Networks}

Originating from \cite{goodfellow2014generative}, generative adversarial networks have undergone numerous modifications and improvements over the past few years \cite{arjovsky2017wasserstein, gulrajani2017improved, zhang2019self, zhu2017unpaired, karras2019style}. In this paper, we focus on quantization of three architecturally distinct variants of GANs, which are CycleGAN \cite{zhu2017unpaired}, SAGAN \cite{zhang2019self} and StyleGAN \cite{karras2019style}. CycleGAN model tackles unpaired image-to-image translation problem by imposing cycle-consistency on the training process. From an architectural point of view, the generator of this model is a fully convolutional neural network with residual blocks. In contrast, the generator of SAGAN is augmented with self-attention layers in order to facilitate long-range dependency modeling. 
Another important type of GANs architecture is a style-based one. It was shown that alteration of generator architecture so that latent codes are provided in adaptive instance normalization blocks instead of the first convolutional layer leads to significant improvements in the quality of generated images as well as increased latent space disentanglement. In this work, we study the impact of quantization on style-based generator characteristics.

\subsection{Quantization of Generative Adversarial Networks}
There have been a few works investigating the quantization of GANs. Wang et al. \cite{wang2019qgan} utilized the expectation-maximization algorithm in order to quantize weights to a non-uniform grid of values while keeping activations unquantized. Despite promising results, the method under consideration cannot be directly used to reduce latency and perform computations in low bitwidth arithmetic; however, one still might find it useful for weights' compression. Wan et al. \cite{wan2020deep} approached the problem of quantization of deep generative models via a learnable non-uniform quantization scheme. Although it is theoretically possible to employ this method for low bitwidth inference in generative models, the non-uniform scheme utilized complicates deployment of this framework on edge devices.  Perhaps, the most relevant method was described in \cite{wang2020gan}. The authors applied uniform quantization to both weights and activations, combining quantization with pruning and knowledge distillation in a unified optimization framework. However, the paper is mainly concerned with compression techniques in general; thus, it misses some important insights about GANs quantization. For instance, the authors considered only 8-bit quantization of fully-convolutional architectures. Moreover, they fixed quantization parameters prior to training, while common practice \cite{esser2019learned} and our experiments show that it is beneficial to tune quantization parameters jointly with weights.

\section{Methodology}

	\label{ch:method}
	
In this section, we provide a description of experimental design choices and motivate them.

\subsection{Uniform and non-uniform quantization}

Generally, two types of quantization are distinguished in a principled way, namely uniform and non-uniform quantization. The first one maps values of weights and activations to a uniform grid of fixed-point representations:
\begin{equation} \label{eq:symquant}
    x_q = \Delta \cdot \mathrm{clamp}\big(\mathrm{round}(\frac{x}{\Delta}), t_{\min}, t_{\max} \big) = \Delta \cdot x_{int},
\end{equation}
where $x$ denotes a full precision real-valued number, $x_q$ is the corresponding quantized value, $\Delta$ is a quantization step size (a parameter to be determined during quantization process), $x_{int}$ is an integer representation, $\mathrm{clamp}(x, t_{\min}, t_{\max})$ denotes clamping $x$ to the $[t_{\min}, t_{\max}]$ interval. By replacing floating-point tensors with uniformly quantized ones, it becomes possible to perform typical calculations (e.g., convolution) in low bitwidth integer arithmetic, thus reducing computational resources. In contrast, non-uniform quantization maps values to a non-equidistant grid of quantized values, thereby being more flexible \cite{zhang2018lq, park2018value}.  However, these approaches are harder to be deployed on hardware, since the execution of low bitwidth operations is not as straightforward as for integer arithmetic in the uniform quantization case. Hence, we consider uniform quantization only.

\subsection{Static and dynamic quantization}

Since the distribution of activations may vary significantly depending on the model input, it is desirable to adapt quantization of activations dynamically during inference. Dynamic quantization does this by recomputing quantization parameters for each batch of data independently. Although dynamic quantization allows for a flexible approximation of activations, it induces considerable overhead in computations. In contrast, static quantization uses precomputed quantization parameters for activations, thus being more computationally efficient. With this being said, we decided to use static quantization to achieve maximal efficiency.

\subsection{Per-chanel and per-tensor weight quantization}

Another important concept is per-channel quantization of weights \cite{krishnamoorthi2018quantizing}. Per-channel quantization assigns distinct parameters (quantization scales) to each convolutional kernel, enabling flexible quantization of weights. At the same time, it does not involve significant overhead in computations. Nevertheless, considerable research efforts were concentrated on eliminating the need for per-channel quantization of weights to simplify the implementation of quantized operations \cite{nagel2019data, nagel2020up}. In our work, we investigate the importance of per-channel quantization for GANs.

\subsection{Symmetric and asymmetric quantization}
The concept of asymmetric (affine) quantization relates to the usage of flexible zero-points in the quantization scheme. Specifically, one can modify the formula \eqref{eq:symquant} in the following way:
\begin{equation} \label{eq:asymquant}
    x_q = \Delta \cdot \bigg ( \mathrm{clamp}\big(\mathrm{round}(\frac{x}{\Delta} + z), t_{\min}, t_{\max} \big) - z \bigg ) = \Delta \cdot (x_{int} - z),
\end{equation}
where $z$ is an integer corresponding to zero point. Hence, asymmetric quantization allows modeling non-zero-centered distributions of weights and activations more effectively. However, it introduces additional computational cost, which is negligible for asymmetric quantization of activations but might be significant in asymmetric quantization of weights \cite{krishnamoorthi2018quantizing}. Thus, for our experiments we use symmetric signed ($t_{\min} = -2^{n-1}$, $t_{\max} = 2^{n-1} - 1$, where $n$ is the number of bits) quantization for weights and asymmetric for activations.

\subsection{Post-training quantization} \label{ptq}

We consider two approaches to post-training quantization of GANs. The first one is based on observation of statistical functions and later referred to as vanilla PTQ. In this approach, we propose to tune quantization parameters so that the minimal and maximal quantized values are matched to the certain quantiles of full-precision values distribution. The optimal quantiles for weights and activations are determined independently and shared between all layers. Despite the simplicity of this approach, we found it to provide significant improvements over naive baseline, which matches minimal and maximal quantized values to minimum and maximum of full-precision values.


Another approach to post-training quantization of GANs is built upon recent works \cite{li2021brecq, esser2019learned, nagel2020up}. Consider a generator model $f$ composed of $n$ sequential blocks $f = f_n \circ  f_{n - 1} \circ \dots \circ f_1$ and its quantized version $f^{q} = f^q_n \circ  f^q_{n - 1} \circ \dots \circ f^q_1$. Let us denote the input and output of the $n$-th full-precision block as $X_n$ and $Y_n$, respectively. Similarly to BRECQ technique \cite{li2021brecq}, we block-wisely optimize weights and quantizers by matching activations' feature maps to the ones of the full-precision model:
\begin{equation} \label{eq:local}
    (\{\Delta_k, z_k\}_k, \{W^q_l\}_l) = \argmin \limits_{\{\Delta_k, z_k\}_k, \{W_l^q\}_l} \| f^q_i(X_{i}) - Y_i \|_F^2,
\end{equation}
where $\{\Delta_k, z_k\}_k$ are the quantization parameters for activations and possibly weights within $f^q_i$, $\{W^q_l\}_l$ are the quantized weights of $f^q_i$. We consider two quantization learning methods. The first one (STE-BRECQ) relies on {straight-through estimator} (STE) \cite{krishnamoorthi2018quantizing} and learned step size quantization (LSQ) \cite{esser2019learned} for differentiation through rounding operation. The second method (AR-BRECQ) employs {adaptive rounding} technique for weights quantization and applies LSQ only for learning quantization of activations (as in \cite{li2021brecq}). We also tried to incorporate squared gradient information into the loss function \eqref{eq:local} as it was proposed in \cite{li2021brecq}, but did not observe any improvements.

\subsection{Quantization-aware training} \label{qat}

Although post-training quantization is fast and requires a small data sample, its performance is restricted by limited resources. Hence, we make use of quantization-aware training to further reduce the quality degradation associated with quantization. Similarly to \cite{wang2020gan}, we employ a sum of adversarial and reconstruction losses for quantization-aware training:

\begin{equation}\label{perceptual}
    L(\phi_q, \theta) = L_{adv} + \beta \cdot L_{rec} = \mathbb{E}_{x \sim \mathrm{inp}} \bigg [ V_{\mathrm{LSGAN}} \big (D(f_q(x; \phi_q); \theta), D(f(x; \phi); \theta) \big ) + \beta \cdot d \big (f_q(x; \phi_q ), f(x; \phi) \big ) \bigg ],
\end{equation}
where $V_{\mathrm{LSGAN}}$ is the least squares GAN loss \cite{mao2017least}, $D(\cdot; \theta)$ denotes the discriminator with its parameters $\theta$, $f(\cdot;\phi)$ and $f_q(\cdot; \phi_q)$ are the full-precision generator and its quantized version, respectively, $d$ denotes the perceptual reconstruction loss \cite{johnson2016perceptual}, $\beta$ is the reconstruction weight hyperparameter. We follow the common practice of using straight-through estimator (STE) \cite{krishnamoorthi2018quantizing} for weights fine-tuning and study two strategies for tuning of quantization parameters. In the first one (MA-QAT), quantization parameters are adjusted by observing statistics of full-precision values during training. Specifically, similarly to vanilla PTQ, the ranges of quantized values are defined by moving averages of observed quantiles.
The quantization parameters are adjusted by such a procedure during the first 50 epochs and remain fixed during subsequent epochs. Another strategy is based on learned step size quantization \cite{esser2019learned} with learnable zero points\footnote{Our implementation is based on codes from PyTorch quantization module (\href{https://github.com/pytorch/pytorch/blob/382781221de4fd3c8035285ce803efb757b5c104/torch/quantization/_learnable_fake_quantize.py}{link}).}. For both strategies, the quantization parameters are initialized using quantiles-based post-training quantization. In addition,  we tried to fix quantization parameters before quantization-aware training, following \cite{wang2020gan}. However, it appeared to be considerably worse than the strategies mentioned above.

\section{Experimental results}
	
	\label{ch:exp}

This section presents experimental results of quantization techniques efficiency assessment for the three GAN models under consideration. We use horse2zebra \cite{zhu2017unpaired}, CelebA \cite{liu2015faceattributes}, and FFHQ (thumbnails 128x128) \cite{karras2019style} datasets to train CycleGAN, SAGAN, and StyleGAN models, respectively. Our experiments are based on the PyTorch quantization module \cite{NEURIPS2019_9015}. We will make the code and checkpoints publicly available. 

It is important to note that in addition to the standard practice of using Frechet inception distance (FID) \cite{heusel2017gans} for comparison of generated images with real ones, we utilize FID to directly measure similarity between images generated by the quantized model and images generated by the full-precision model (using the same latent codes). We found this metric, which we refer to as qFID, to be a more informative reconstruction quality measure.

\subsection{Post-training quantization}

\begin{figure}%
    \centering
    \subfloat[\centering Vanilla PTQ]{{\includegraphics[width=7.5cm]{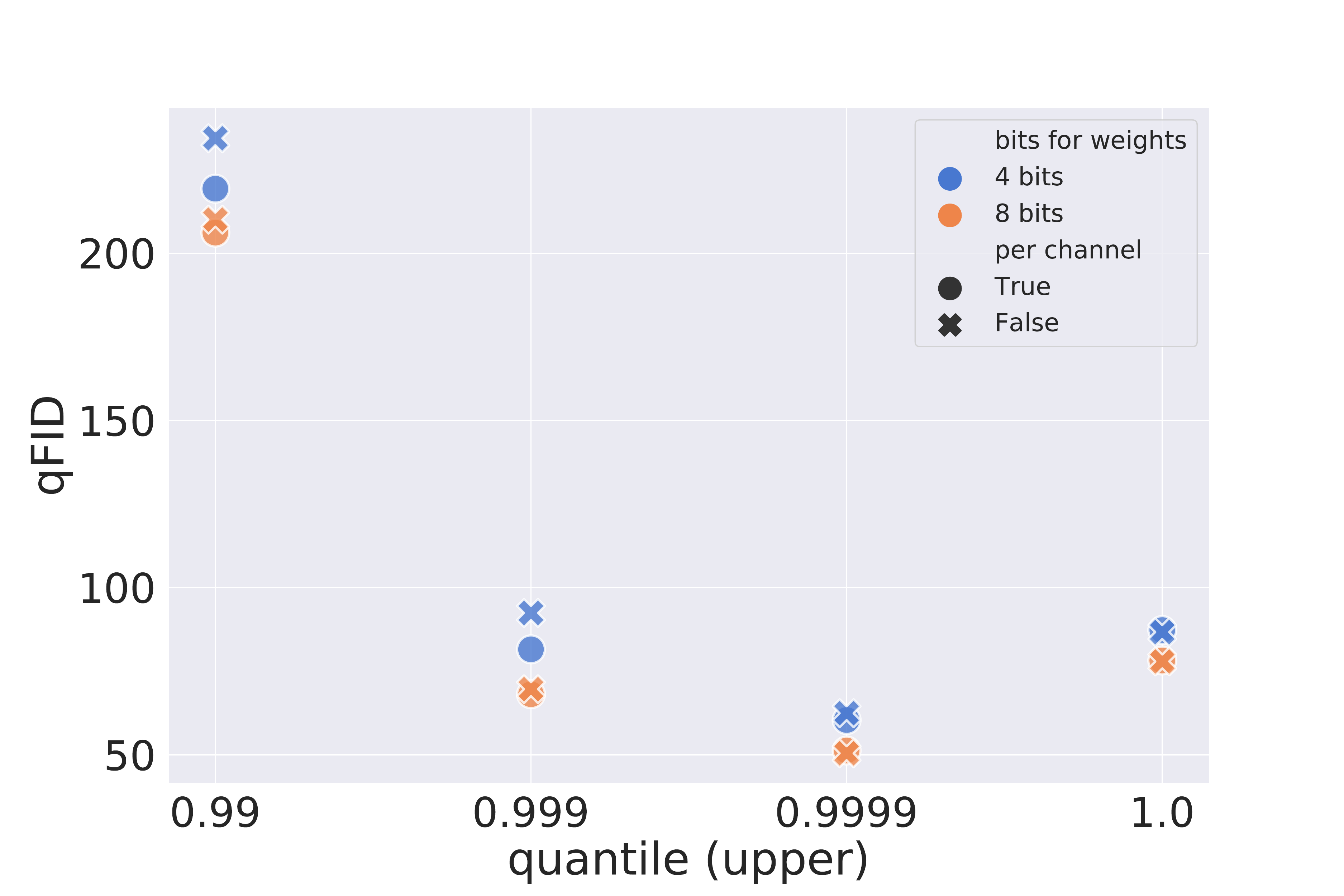} }}%
    \qquad
    \subfloat[\centering MA-QAT (a8/w8, per-channel)]{{\includegraphics[width=7.5cm]{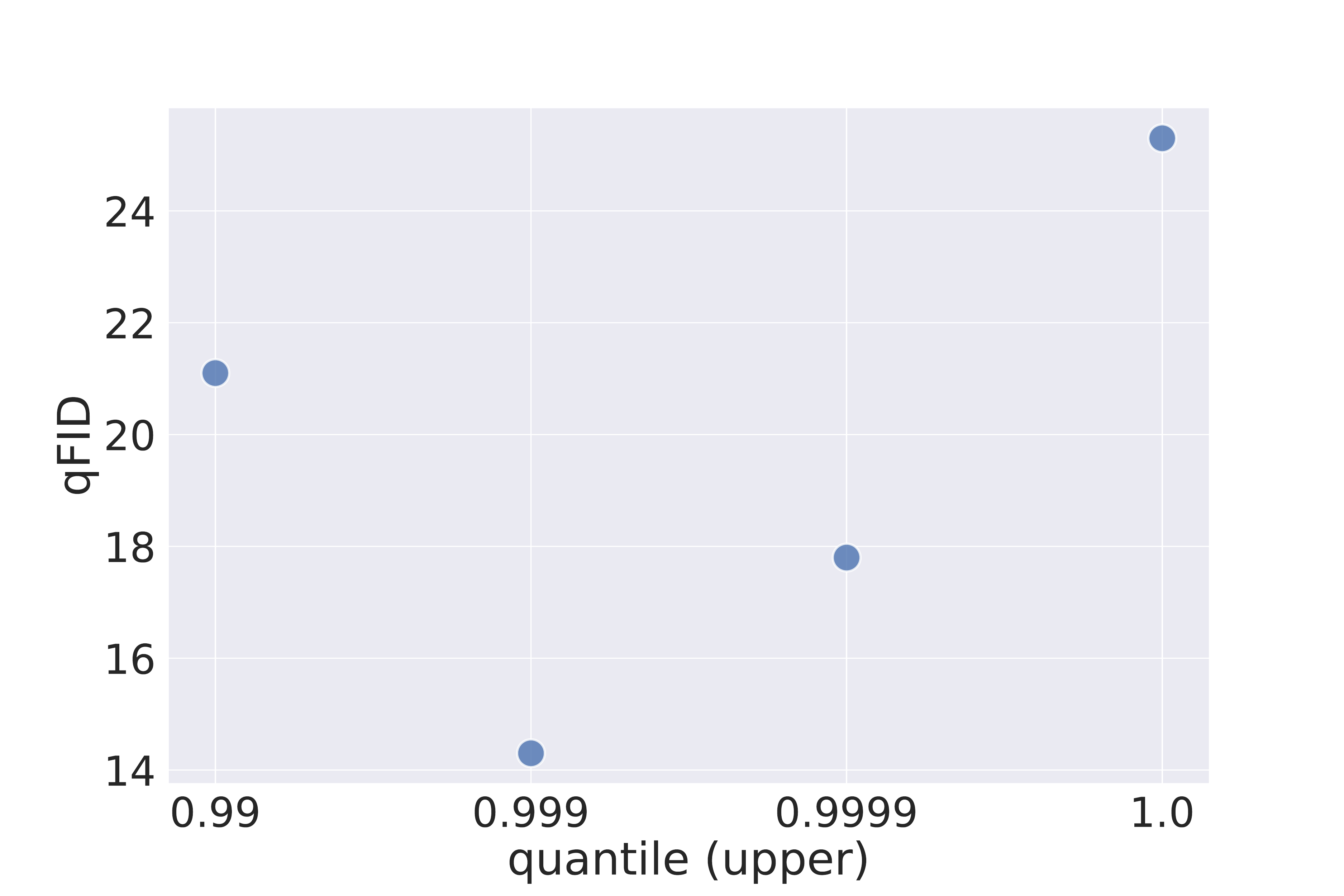} }}%
    \caption{Dependency of StyleGAN qFID on quantile corresponding to the upper bound of quantized activations range (the lower bound quantile is chosen symmetrically). }%
    \label{fig:quantiles}%
\end{figure}

Vanilla post-training quantization considered in this work is based on observation of quantiles of weights' and activations' distributions. Specifically, we tune the quantization ranges by a moving average of observed quantiles during the generation of 500 images (one by one), with moving average momentum being equal to 0.99. We examine 1.0/0.0, 0.9999/0.0001, 0.999/0.001, 0.99/0.01 pairs of quantiles for upper and lower quantization range bounds, respectively, and observe significant dependency of quality of generated images on quantiles' values for activations (see Figure \ref{fig:quantiles}). In contrast, no significant improvements were observed during tuning quantiles for weights. Based on our experiments, we find pairs of quantiles 0.9999/0.0001 to be optimal for post-training quantization of activations for StyleGAN and CycleGAN, whereas pair 1.0/0.0 is optimal for SAGAN. For weights' quantization, we observe the pair of quantiles 0.9999/0.0001 to perform slightly better than other tested configurations and use it throughout our experiments.

\begin{figure}[h!]
\begin{center}
\centerline{\includegraphics[width=1.0\columnwidth]{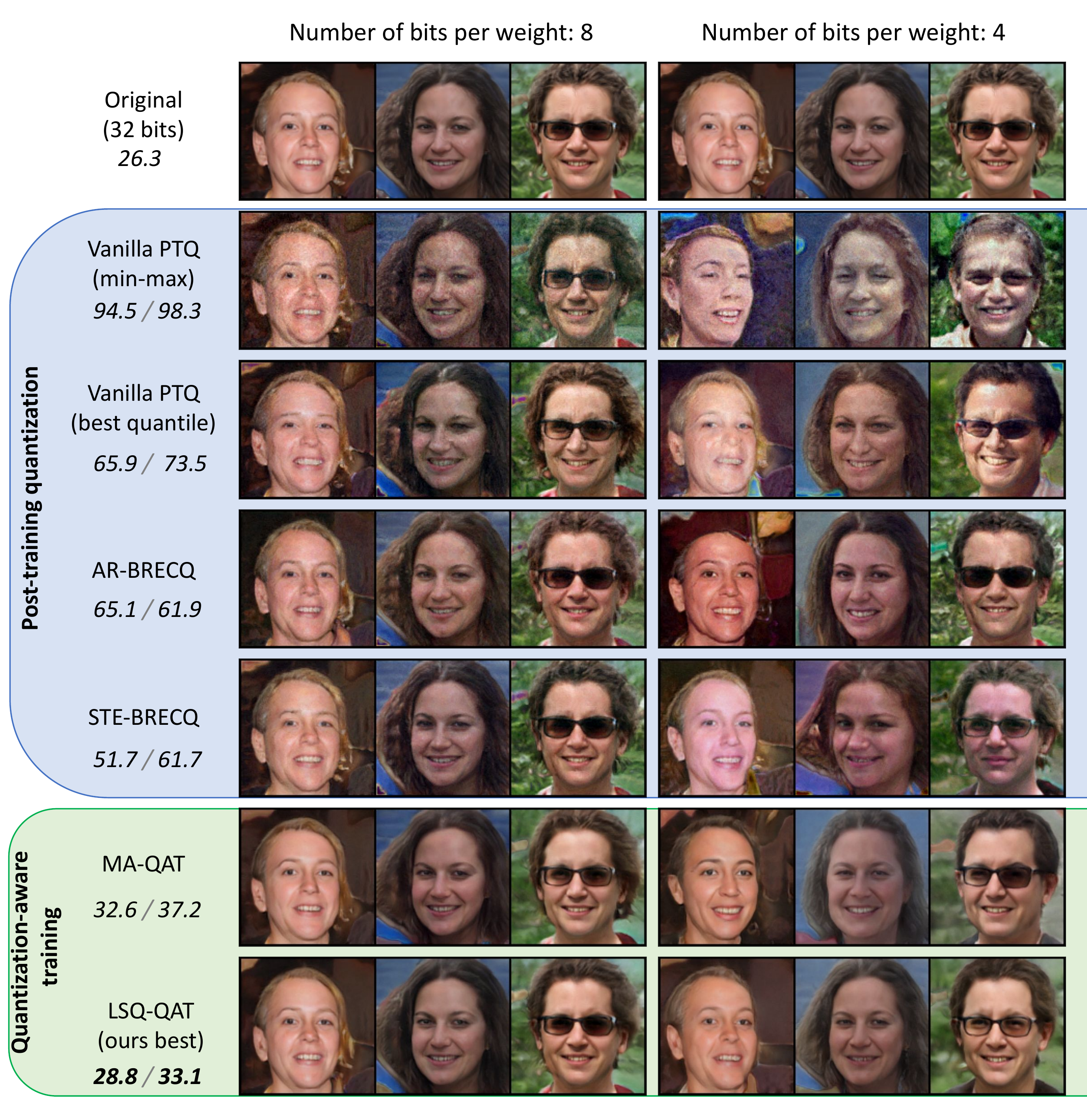}}
\caption{Sample images along with Frechet inception distance scores~(lower is better, $\cdot$ \textcolor{gray}{/} $\cdot$ correspond to 4\textcolor{gray}{/}8-bit weights respectively) for  various quantization methods applied to StyleGAN generator. 
Vanilla PTQ with quantiles-based statistics matching (quantile) generates significantly better samples than baseline (min-max).
In general, quantization-aware training~(highlighted by \textcolor{caribbeangreen}{green}) produces significantly more realistic samples than post-training quantization~(highlighted by \textcolor{bluebell}{blue}). 
The learned step size quantization with quantile initialization~(LSQ-QAT) allows quantization with negligible quality degradation. Activations are quantized to 8 bit, weights are quantized per-tensor with the number of bits outlined at the top of the figure. Note that for the generation of these images we have used the truncation trick to avoid sampling from extreme regions \cite{karras2019style}.}
\label{fig:sample}
\end{center}
\end{figure}

In addition to vanilla post-training quantization, we explore advanced PTQ techniques described in \ref{ptq} for style-based generator architecture. We consider styled convolutional blocks as blocks for the reconstruction of synthesis network and pairs of linear layer and subsequent nonlinearity (leaky ReLU) for the reconstruction of mapping network (see \cite{karras2019style} for definition mapping and synthesis networks).  
The results of the experiments are summarized in Table \ref{table:ptq}. For both strategies, we use a training sample of size 896 images and batch size of 16 images; we refer the reader to the Appendix \ref{app:impl} for other implementation details. An interesting finding of this work is that STE-based BRECQ significantly outperforms AdaRound-based BRECQ for 8-bit quantization while being inferior for 4-bit quantization. Furthermore, AR-BRECQ provides better FID for 4-bit weights quantization than for 8-bit. We attribute this counterintuitive result to the nature of adaptive rounding, which by design allows only for local optimization of rounding rule. Thus, the 4-bit adaptive rounding may have richer space for optimization than the 8-bit one due to larger quantization step sizes. Overall, the results suggest that post-training quantization may provide samples of reasonable quality (see Figure \ref{fig:sample}); however, to go further, quantization-aware training is needed.

\begin{table}[!h]
  \caption{Performance (FID and qFID, lower is better) of post-training quantization techniques in application to quantization of style-based generator. }   \label{table:ptq}
    \vskip 0.15in
 \begin{center}
   \begin{tabular}{|l|c|>{\centering\arraybackslash}p{1.8cm}>{\centering\arraybackslash}p{1.8cm}|>{\centering\arraybackslash}p{1.8cm}>{\centering\arraybackslash}p{1.8cm}|}
 \hline 
    \multirow{2}{*}{Method} & \multirow{2}{*}{Per-channel}  &
      \multicolumn{2}{c|}{Number of bits per weight: 8} &
       \multicolumn{2}{c|}{Number of bits per weight: 4}  \\
      & & {FID ($\downarrow$)} & {qFID ($\downarrow$)} & {FID ($\downarrow$)} & {qFID ($\downarrow$)}  \\
      \hline
    full-precision & - & 26.3 & 0.0  & 26.3 & 0.0   \\ \hline
     \multirow{2}{*}{vanilla PTQ} & \checkmark & 66.5  & 51.2 & 73.1 & 60.3  \\
     & $\times$  & 65.9 & 50.4 & 73.5 & 62.4  \\ \hline
      
     \multirow{2}{*}{AR-BRECQ} & \checkmark  & 66.8 & 49.9  & \textbf{61.2} & \textbf{46.2} \\
     & $\times$ & 65.1 & 48.4 & 61.9 & 48.1  \\ \hline
    
    STE-BRECQ  & $\times$ &  \textbf{51.7} & \textbf{35.9} & 61.7 & 51.6  \\

    \hline
  \end{tabular}
   \end{center}

\end{table}


\subsection{Quantization-aware training}

As described in \ref{qat}, for quantization-aware training, we use a sum of adversarial and perceptual losses. Thus, the resulting loss function has three additive components, namely adversarial loss, perceptual content loss, and perceptual style loss. Each term's relative weights are tuned via the grid search over 27 parameter combinations to optimize performance on the SAGAN quantization. The search space is a cartesian product of the following sets $\{0, 0.01, 0.1\}$ (adversarial loss), $\{1, 3, 10\}$ (content loss) and $\{10^4, 3 \cdot 10^4, 10^5\}$ (style loss). We did not find huge deviations in the results depending on the combination used. However, one combination performed slightly better than others, that is $0.01$ coefficient for adversarial loss, $3$ for content loss, and $3 \cdot 10^4$ for style loss. Given that these parameters did not account for a significant impact on the performance, we did not re-run the grid search for other concerned GAN models and used this set of parameters for all quantization-aware training experiments. Besides, we studied the importance of the considered distillation objective by replacing it with pixel-wise mean squared error (MSE). Overall, the perceptual loss significantly outperformed the MSE loss for the concerned models; please refer to the Appendix \ref{app:obj} for detailed results.

The LSQ-based procedure for tuning quantization parameters appears to outperform the moving average-based one considerably (see Table \ref{Tab:48bit_table}). This effect is especially noticeable in the case of 4-bit quantization of weights. It is noteworthy that we found it beneficial for the CycleGAN model to perform 50 initial training epochs using the MA-QAT procedure while applying the LSQ-based one during the subsequent 50 epochs. Besides, the optimal quantiles for MA-QAT depend on a particular GAN model and do not necessarily coincide with those selected at the post-training quantization stage (see Figure \ref{fig:quantiles}). We found the pairs 0.999/0.001, 0.9999/0.0001, and 0.999/0.001 to give optimal performance for StyleGAN, SAGAN, and Cycle-GAN, respectively.

\begin{table}[!h]
  \caption{Performance of quantization-aware training techniques. MA-QAT and LSQ-QAT denote moving average and learned step size strategies for tuning quantization parameters. “wxay” means x-bit weights
and y-bit activations. $^*$: LSQ is applied only during the last 50 epochs.} \label{48bit_table} 
    \vskip 0.15in
 \centering
  \begin{tabular}{|l|c|c|>{\centering\arraybackslash}p{1.3cm}>{\centering\arraybackslash}p{1.3cm}|>{\centering\arraybackslash}p{1.3cm}>{\centering\arraybackslash}p{1.3cm}|>{\centering\arraybackslash}p{1.3cm}>{\centering\arraybackslash}p{1.3cm}|}
 \hline 
     \multirow{2}{*}{Method} & \multirow{2}{*}{Per-channel} & \multirow{2}{*}{Bits}  &
      \multicolumn{2}{c|}{CycleGAN} &
       \multicolumn{2}{c|}{SAGAN} &
     \multicolumn{2}{c|}{StyleGAN}  \\
      & & & {FID ($\downarrow$)} & {qFID ($\downarrow$)} & {FID ($\downarrow$)} & {qFID ($\downarrow$)} & {FID ($\downarrow$)} & {qFID ($\downarrow$)} \\
      \hline
    full-precision & - & - & 62.4 & 0.0 & 138.6  & 0.0 & 26.3 & 0.0 \\ \hline
     \multirow{4}{*}{vanilla PTQ} & \checkmark & a8/w8  & 98.8  & 71.9 & 139.6 & 81.1 & 66.5 & 51.2 \\
     & $\times$ & a8/w8  & 103.0 & 74.5 & 145.3 & 86.8 & 65.9 & 50.4 \\
     & \checkmark & a8/w4  & 118.3 &  85.1 & 265.4 & 287.2 & 73.1 & 60.3 \\
     &  $\times$ &  a8/w4 & 154.0  & 115.5 & 309.4 & 325.2 & 73.5 & 62.4 \\ \hline

     \multirow{4}{*}{MA-QAT} & \checkmark & a8/w8  & 63.7 & 38.3  & 125.9 & 10.4 & 32.3 & 14.3 \\
     & $\times$ & a8/w8  & \textbf{63.1} & 40.8 & \textbf{124.8} & 11.9 & 32.6 & 14.7 \\
     & \checkmark & a8/w4  & \textbf{69.3} & 50.7 & \textbf{114.2} & 29.1  & 40.1 & 23.3 \\
     &  $\times$ &  a8/w4  & 112.5 & 88.3 & 128.7 & 31.5 & 37.2 & 18.5 \\ \hline
      
    
   \multirow{2}{*}{LSQ-QAT }  & $\times$ & a8/w8  & \textbf{63.1}$^*$ & \textbf{36.1}$^{*}$ & 130.1 & \textbf{5.6} & \textbf{28.8} & \textbf{8.5} \\
     &  $\times$ &  a8/w4  & 71.4$^{*}$ & \textbf{47.6}$^{*}$ & 125.8 & \textbf{9.0} & \textbf{33.1} & \textbf{11.0}\\ 

    \hline
  \end{tabular}

 \label{Tab:48bit_table}
\end{table}

\section{Discussion}

	\label{ch:disc}

The results of experiments demonstrate that GANs \underline{can be} successfully quantized for fixed-point inference with negligible quality degradation (see Figure \ref{fig:sample}).
In particular, we show that \emph{learned step size quantization} provides a recipe for the successful quantization of generative adversarial networks. We find this approach to generalize across the most important GAN architectures, specifically StyleGAN, SAGAN, and CycleGAN.

Interestingly, we reveal that, in contrast to classification models, GANs cannot be quantized with modern methods of post-training quantization and require more resource-demanding methods of quantization-aware training. 
The results also shed light on what technical details matter for successful GAN quantization. 
For instance, we show that using quantiles instead of minimal and maximal values for tuning quantization intervals is crucial for well-performing post-training quantization and moving average-based quantization-aware training. 
In accordance with prior work \cite{krishnamoorthi2018quantizing}, we show that per-channel quantization does not provide significant benefits when using quantization-aware training.

\section{Conclusion}

	\label{ch:disc}

In this work, we have conducted an extensive experimental study of state-of-art quantization techniques' effectiveness to uniform quantization of generative adversarial networks. The study concerns post-training quantization and quantization-aware training techniques applied to generators of three diverse GAN architectures, namely style-based GAN, self-attention GAN, and cycle-consistent GAN. The results suggest that these models can be successfully quantized for 4/8-bit inference with negligible quality degradation. We hope that the results of this paper will set up a starting point for future research on the quantization of GANs and allow using GANs in a lot of new AI-driven products.

\section{Acknowledgments}

	
This work was supported by Samsung Research. We also thank Egor Zakharov for helpful comments.

\newpage
\printbibliography

@inproceedings{nagel2019data,
  title={Data-free quantization through weight equalization and bias correction},
  author={Nagel, Markus and Baalen, Mart van and Blankevoort, Tijmen and Welling, Max},
  booktitle={Proceedings of the IEEE International Conference on Computer Vision},
  pages={1325--1334},
  year={2019}
}

@article{nagel2020up,
  title={Up or Down? Adaptive Rounding for Post-Training Quantization},
  author={Nagel, Markus and Amjad, Rana Ali and van Baalen, Mart and Louizos, Christos and Blankevoort, Tijmen},
  journal={arXiv preprint arXiv:2004.10568},
  year={2020}
}

@article{yao2020hawqv3,
  title={HAWQV3: Dyadic Neural Network Quantization},
  author={Yao, Zhewei and Dong, Zhen and Zheng, Zhangcheng and Gholami, Amir and Yu, Jiali and Tan, Eric and Wang, Leyuan and Huang, Qijing and Wang, Yida and Mahoney, Michael W and others},
  journal={arXiv preprint arXiv:2011.10680},
  year={2020}
}

@article{esser2019learned,
  title={Learned step size quantization},
  author={Esser, Steven K and McKinstry, Jeffrey L and Bablani, Deepika and Appuswamy, Rathinakumar and Modha, Dharmendra S},
  journal={arXiv preprint arXiv:1902.08153},
  year={2019}
}

@article{jain2019trained,
  title={Trained quantization thresholds for accurate and efficient fixed-point inference of deep neural networks},
  author={Jain, Sambhav R and Gural, Albert and Wu, Michael and Dick, Chris H},
  journal={arXiv preprint arXiv:1903.08066},
  year={2019}
}

@article{krishnamoorthi2018quantizing,
  title={Quantizing deep convolutional networks for efficient inference: A whitepaper},
  author={Krishnamoorthi, Raghuraman},
  journal={arXiv preprint arXiv:1806.08342},
  year={2018}
}

@article{hubara2020improving,
  title={Improving post training neural quantization: Layer-wise calibration and integer programming},
  author={Hubara, Itay and Nahshan, Yury and Hanani, Yair and Banner, Ron and Soudry, Daniel},
  journal={arXiv preprint arXiv:2006.10518},
  year={2020}
}

@inproceedings{wang2020towards,
  title={Towards accurate post-training network quantization via bit-split and stitching},
  author={Wang, Peisong and Chen, Qiang and He, Xiangyu and Cheng, Jian},
  booktitle={International Conference on Machine Learning},
  pages={9847--9856},
  year={2020},
  organization={PMLR}
}

@inproceedings{zhang2018lq,
  title={Lq-nets: Learned quantization for highly accurate and compact deep neural networks},
  author={Zhang, Dongqing and Yang, Jiaolong and Ye, Dongqiangzi and Hua, Gang},
  booktitle={Proceedings of the European conference on computer vision (ECCV)},
  pages={365--382},
  year={2018}
}

@article{wang2019qgan,
  title={QGAN: Quantized generative adversarial networks},
  author={Wang, Peiqi and Wang, Dongsheng and Ji, Yu and Xie, Xinfeng and Song, Haoxuan and Liu, XuXin and Lyu, Yongqiang and Xie, Yuan},
  journal={arXiv preprint arXiv:1901.08263},
  year={2019}
}

@article{wan2020deep,
  title={Deep quantization generative networks},
  author={Wan, Diwen and Shen, Fumin and Liu, Li and Zhu, Fan and Huang, Lei and Yu, Mengyang and Shen, Heng Tao and Shao, Ling},
  journal={Pattern Recognition},
  volume={105},
  pages={107338},
  year={2020},
  publisher={Elsevier}
}

@inproceedings{wang2020gan,
  title={GAN Slimming: All-in-One GAN Compression by A Unified Optimization Framework},
  author={Wang, Haotao and Gui, Shupeng and Yang, Haichuan and Liu, Ji and Wang, Zhangyang},
  booktitle={European Conference on Computer Vision},
  pages={54--73},
  year={2020},
  organization={Springer}
}

@inproceedings{karras2019style,
  title={A style-based generator architecture for generative adversarial networks},
  author={Karras, Tero and Laine, Samuli and Aila, Timo},
  booktitle={Proceedings of the IEEE/CVF Conference on Computer Vision and Pattern Recognition},
  pages={4401--4410},
  year={2019}
}

@inproceedings{zhu2017unpaired,
  title={Unpaired image-to-image translation using cycle-consistent adversarial networks},
  author={Zhu, Jun-Yan and Park, Taesung and Isola, Phillip and Efros, Alexei A},
  booktitle={Proceedings of the IEEE international conference on computer vision},
  pages={2223--2232},
  year={2017}
}

@article{zhang2017beyond,
  title={Beyond a gaussian denoiser: Residual learning of deep cnn for image denoising},
  author={Zhang, Kai and Zuo, Wangmeng and Chen, Yunjin and Meng, Deyu and Zhang, Lei},
  journal={IEEE transactions on image processing},
  volume={26},
  number={7},
  pages={3142--3155},
  year={2017},
  publisher={IEEE}
}

@article{brock2018large,
  title={Large scale GAN training for high fidelity natural image synthesis},
  author={Brock, Andrew and Donahue, Jeff and Simonyan, Karen},
  journal={arXiv preprint arXiv:1809.11096},
  year={2018}
}

@inproceedings{zhang2019self,
  title={Self-attention generative adversarial networks},
  author={Zhang, Han and Goodfellow, Ian and Metaxas, Dimitris and Odena, Augustus},
  booktitle={International conference on machine learning},
  pages={7354--7363},
  year={2019},
  organization={PMLR}
}

@article{goodfellow2014generative,
  title={Generative adversarial networks},
  author={Goodfellow, Ian J and Pouget-Abadie, Jean and Mirza, Mehdi and Xu, Bing and Warde-Farley, David and Ozair, Sherjil and Courville, Aaron and Bengio, Yoshua},
  journal={arXiv preprint arXiv:1406.2661},
  year={2014}
}

@inproceedings{arjovsky2017wasserstein,
  title={Wasserstein generative adversarial networks},
  author={Arjovsky, Martin and Chintala, Soumith and Bottou, L{\'e}on},
  booktitle={International conference on machine learning},
  pages={214--223},
  year={2017},
  organization={PMLR}
}

@article{gulrajani2017improved,
  title={Improved training of wasserstein gans},
  author={Gulrajani, Ishaan and Ahmed, Faruk and Arjovsky, Martin and Dumoulin, Vincent and Courville, Aaron},
  journal={arXiv preprint arXiv:1704.00028},
  year={2017}
}

@article{heusel2017gans,
  title={Gans trained by a two time-scale update rule converge to a local nash equilibrium},
  author={Heusel, Martin and Ramsauer, Hubert and Unterthiner, Thomas and Nessler, Bernhard and Hochreiter, Sepp},
  journal={arXiv preprint arXiv:1706.08500},
  year={2017}
}

@incollection{NEURIPS2019_9015,
title = {PyTorch: An Imperative Style, High-Performance Deep Learning Library},
author = {Paszke, Adam and Gross, Sam and Massa, Francisco and Lerer, Adam and Bradbury, James and Chanan, Gregory and Killeen, Trevor and Lin, Zeming and Gimelshein, Natalia and Antiga, Luca and Desmaison, Alban and Kopf, Andreas and Yang, Edward and DeVito, Zachary and Raison, Martin and Tejani, Alykhan and Chilamkurthy, Sasank and Steiner, Benoit and Fang, Lu and Bai, Junjie and Chintala, Soumith},
booktitle = {Advances in Neural Information Processing Systems 32},
editor = {H. Wallach and H. Larochelle and A. Beygelzimer and F. d\textquotesingle Alch\'{e}-Buc and E. Fox and R. Garnett},
pages = {8024--8035},
year = {2019},
publisher = {Curran Associates, Inc.},
url = {http://papers.neurips.cc/paper/9015-pytorch-an-imperative-style-high-performance-deep-learning-library.pdf}
}

@article{li2021brecq,
  title={BRECQ: Pushing the Limit of Post-Training Quantization by Block Reconstruction},
  author={Li, Yuhang and Gong, Ruihao and Tan, Xu and Yang, Yang and Hu, Peng and Zhang, Qi and Yu, Fengwei and Wang, Wei and Gu, Shi},
  journal={arXiv preprint arXiv:2102.05426},
  year={2021}
}

@inproceedings{mao2017least,
  title={Least squares generative adversarial networks},
  author={Mao, Xudong and Li, Qing and Xie, Haoran and Lau, Raymond YK and Wang, Zhen and Paul Smolley, Stephen},
  booktitle={Proceedings of the IEEE international conference on computer vision},
  pages={2794--2802},
  year={2017}
}

@inproceedings{johnson2016perceptual,
  title={Perceptual losses for real-time style transfer and super-resolution},
  author={Johnson, Justin and Alahi, Alexandre and Fei-Fei, Li},
  booktitle={European conference on computer vision},
  pages={694--711},
  year={2016},
  organization={Springer}
}

@inproceedings{molchanov2017variational,
  title={Variational dropout sparsifies deep neural networks},
  author={Molchanov, Dmitry and Ashukha, Arsenii and Vetrov, Dmitry},
  booktitle={International Conference on Machine Learning},
  pages={2498--2507},
  year={2017},
  organization={PMLR}
}

@inproceedings{he2017channel,
  title={Channel pruning for accelerating very deep neural networks},
  author={He, Yihui and Zhang, Xiangyu and Sun, Jian},
  booktitle={Proceedings of the IEEE International Conference on Computer Vision},
  pages={1389--1397},
  year={2017}
}

@inproceedings{chollet2017xception,
  title={Xception: Deep learning with depthwise separable convolutions},
  author={Chollet, Fran{\c{c}}ois},
  booktitle={Proceedings of the IEEE conference on computer vision and pattern recognition},
  pages={1251--1258},
  year={2017}
}

@article{howard2017mobilenets,
  title={Mobilenets: Efficient convolutional neural networks for mobile vision applications},
  author={Howard, Andrew G and Zhu, Menglong and Chen, Bo and Kalenichenko, Dmitry and Wang, Weijun and Weyand, Tobias and Andreetto, Marco and Adam, Hartwig},
  journal={arXiv preprint arXiv:1704.04861},
  year={2017}
}

@article{hinton2015distilling,
  title={Distilling the knowledge in a neural network},
  author={Hinton, Geoffrey and Vinyals, Oriol and Dean, Jeff},
  journal={arXiv preprint arXiv:1503.02531},
  year={2015}
}

@inproceedings{chen2017learning,
  title={Learning efficient object detection models with knowledge distillation},
  author={Chen, Guobin and Choi, Wongun and Yu, Xiang and Han, Tony and Chandraker, Manmohan},
  booktitle={Proceedings of the 31st International Conference on Neural Information Processing Systems},
  pages={742--751},
  year={2017}
}

@inproceedings{park2018value,
  title={Value-aware quantization for training and inference of neural networks},
  author={Park, Eunhyeok and Yoo, Sungjoo and Vajda, Peter},
  booktitle={Proceedings of the European Conference on Computer Vision (ECCV)},
  pages={580--595},
  year={2018}
}

@inproceedings{liu2015faceattributes,
 title = {Deep Learning Face Attributes in the Wild},
 author = {Liu, Ziwei and Luo, Ping and Wang, Xiaogang and Tang, Xiaoou},
 booktitle = {Proceedings of International Conference on Computer Vision (ICCV)},
 month = {December},
 year = {2015} 
}

\newpage
\appendix

	\label{ch:app}

\section{Implementation details} \label{app:impl}

Following common practice, we artificially simulate fixed-point inference by quantizing weights and intermediate activations to fixed-point representations while performing all computations in full-precision arithmetic. All simulations are based on PyTorch fake quantization interface. The experiments are performed using NVIDIA Tesla GPUs.

\subsection{StyleGAN}

The experiments with style-based generative-adversarial networks are based on non-official publicly available implementation accessible by this \href{https://github.com/rosinality/style-based-gan-pytorch}{link}. We make use of the model trained on the FFHQ dataset to produce images in 128x128 resolution. All Freshet inception distance computations are performed on sets of $50 000$ randomly sampled images (the real ones are sampled from thumbnails available at the dataset \href{https://github.com/NVlabs/ffhq-dataset}{repository}, whereas generated ones are obtained by sampling from the latent space without truncation trick). 

Quantization-aware training experiments are conducted on a pre-generated distillation dataset containing $10000$ images produced by a full-precision model and corresponding latent space samples. We found this number of images to be enough for quantization-aware training and did not observe any signs of overfitting. We use Adam optimizer with $\beta_1=0.5$, $\beta_2=0.999$ and  learning rate equal to $10^{-5}$ for weights and $10^{-6}$ for quantization parameters (in case LSQ is used), batch size is equal to 8. The learning rate for weights was selected among $10^{-3}$, $10^{-4}$ and $10^{-5}$, while learning rate quantization parameters was chosen from $10^{-5}$, $3 \cdot 10^{-6}$, $10^{-6}$, $10^{-7}$ and 0. In addition, we tried to use SGD optimizer instead of Adam, as it was described in the original LSQ paper, but that found Adam's performance to be more stable. We fix the momentum of exponential moving average updates to 0.99 for MA-QAT experiments. The model is trained for 200 epochs with the cosine annealing learning rate schedule.

Our AdaRound implementation is based on this \href{https://github.com/yhhhli/BRECQ/blob/main/quant/adaptive_rounding.py}{code} provided by the authors of BRECQ. We use Adam optimizer with $\beta_1=0.9$, $\beta_2=0.999$, learning rate is selected among $10^{-2}, 10^{-3}$ for weights and biases, and among $10^{-5}, 10^{-6}, 10^{-7}$ for quantization parameters (LSQ), batch size is equal to 16. In AdaRound-based BRECQ each block is optimized for 20000 steps with 4000 steps of initial warmup period during which rounding regularization is not applied. For STE-based BRECQ we found 2000 steps to be enough for convergence and use this number of steps throughout STE-BRECQ experiments. Likewise BRECQ, we also tried to include squared gradient information into the reconstruction objective. We employ mean squared error between final layer outputs of the quantized model and full-precision one to compute the layerwise activations' gradients. However, this approach did not account for any improvements compared to simple per-layer MSE-based reconstruction.

\subsection{SAGAN}

We use implementation of the self-attention generative adversarial network accessible by this \href{https://github.com/heykeetae/Self-Attention-GAN}{link}. The model architecture is adapted to be compatible with the PyTorch quantization interface (transposed convolutions are replaced with Upsample+Convolution blocks). The resulting model is trained on CelebA dataset following the implementation concerned. The only significant difference compared to the StyleGAN case is that self-attention layers make the LSQ procedure less stable, so we use a lower learning rate of $10^{-7}$ for quantization parameters. We also increase the batch size to 64. The FID computation procedure is the same as the one utilized for StyleGAN.

\subsection{CycleGAN}

The code for the cycle-consistent generative adversarial network is adopted from the repository accessible by \href{https://github.com/VITA-Group/GAN-Slimming}{link}. For our experiments, we consider the horse-to-zebra dataset. The model is trained and distilled on the train part of the dataset, while FIDs and qFIDs are computed on the test part. Despite a small training dataset (around 1000 images), we still do not observe any overfitting in this case. Similarly to StyleGAN and SAGAN we use Adam optimizer with a learning rate being equal to $10^{-5}$ for weights. For quantization parameters, we use a learning rate of $10^{-6}$. The batch size is equal to 8 and the model is trained for 100 epochs.

\section{Objective importance} \label{app:obj}

We assessed the importance of the perceptual distillation objective by replacing it with pixel-wise mean squared error loss for moving average-based quantization-aware training. The experiments were run with 8-bit per-channel quantization for different pairs of activations quantiles (best result is reported) for StyleGAN and SAGAN models. The resulting qFIDs are 33.2/10.4 for MSE/perceptual objectives for the SAGAN model and 41.2/14.3 for the StyleGAN model. Thus, we validated the importance of perceptual loss for quantization-aware training.

\section{Additional samples}
\label{app:samles}
\begin{figure*}[h!]
\begin{center}
\centerline{\includegraphics[width=0.9\columnwidth]{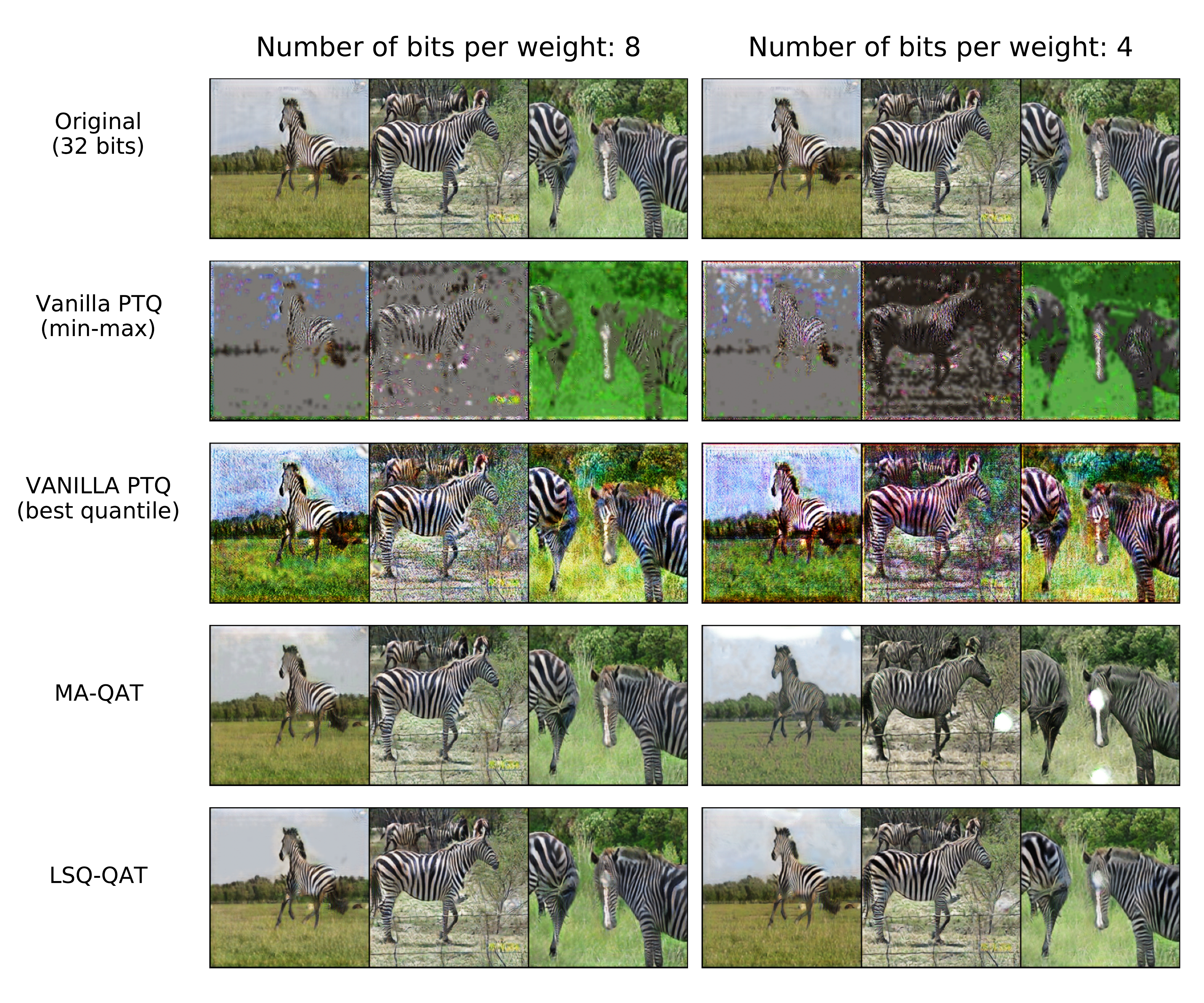}}
\caption{Sample images for  various quantization methods applied to CycleGAN generator. Activations are quantized to 8 bit, weights are quantized per-tensor with the number of bits outlined at the top of the figure.}
\label{sample_cycle}
\end{center}
\vskip -0.2in
\end{figure*}

\begin{figure*}[h!]
\begin{center}
\centerline{\includegraphics[width=0.9\columnwidth]{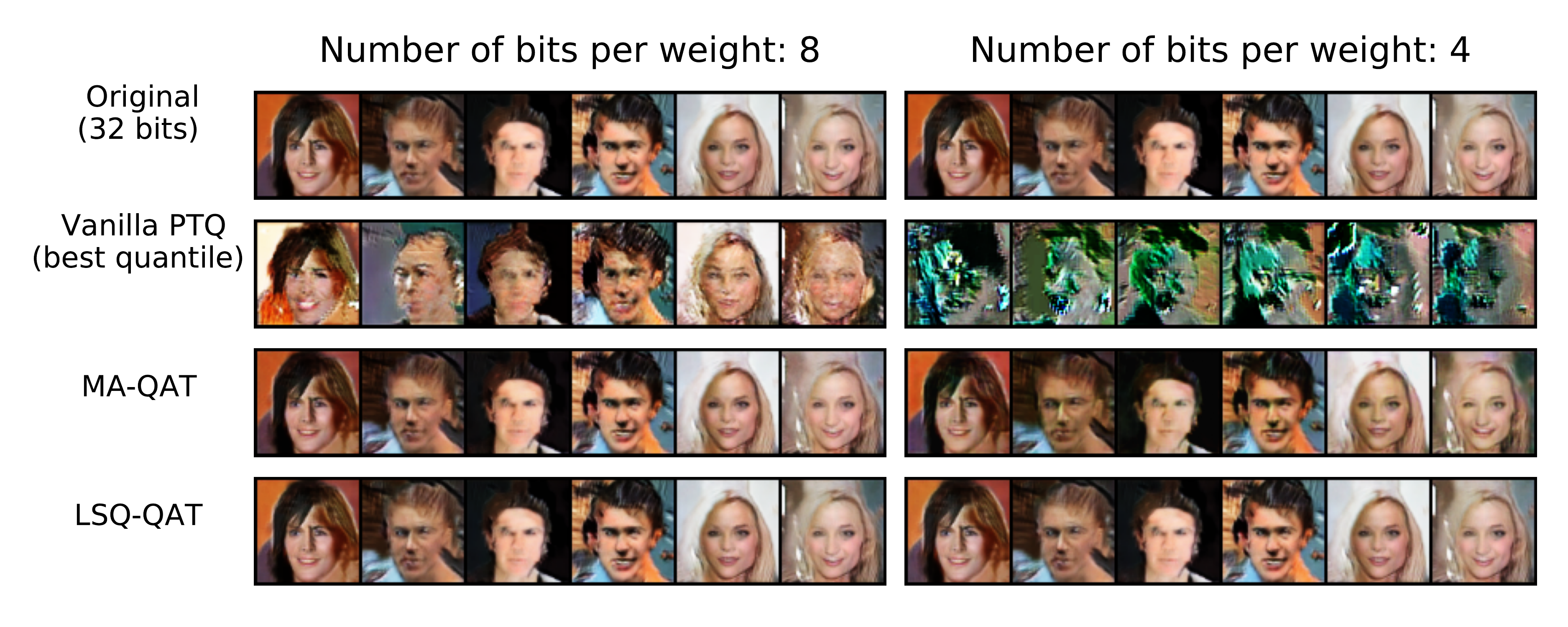}}
\caption{Sample images for  various quantization methods applied to SAGAN generator. Activations are quantized to 8 bit, weights are quantized per-tensor with the number of bits outlined at the top of the figure.}
\label{sample_sagan}
\end{center}
\vskip -0.2in
\end{figure*}

We provide samples for CycleGAN and SAGAN models in Figures \ref{sample_cycle} and \ref{sample_sagan}, respectively. One may refer to Table 2 in the main text for corresponding FIDs and qFIDs.
\vspace{-0.3cm}
\vspace{-0.3cm}

\end{document}